\documentclass[letterpaper, 10 pt, conference]{ieeeconf}  

\IEEEoverridecommandlockouts                              

\usepackage{graphicx}
\usepackage{algorithm}
\usepackage{algorithmic}
\usepackage{mathrsfs}
\usepackage{amsmath}
\usepackage{amssymb}
\usepackage{amsfonts}
\usepackage[colorinlistoftodos]{todonotes}
\usepackage{multirow}
\usepackage{subfig}
\usepackage{bbm}
\title{\LARGE \bf
Active Object Perceiver: Recognition-guided Policy Learning for Object Searching on Mobile Robots
}

\author{Xin Ye$^{1}$, Zhe Lin$^{2}$, Haoxiang Li$^{3}$, Shibin Zheng$^{1}$, Yezhou Yang$^{1}$
                \thanks{$^{1}$ Xin Ye, Shibin Zheng and Yezhou Yang are with the Active Perception Group at the School of Computing, Informatics, and Decision Systems Engineering, Arizona State University, Tempe, AZ, USA, Email:
        {\tt \small  \{xinye1, szheng31, yz.yang\}@asu.edu}}
         \thanks{$^{2}$ Zhe Lin is with Adobe Systems, Inc. San Jose, CA, USA, Email:
        {\tt \small  zlin@adobe.com}}
         \thanks{$^{3}$ Haoxiang Li is with Aibee. Email:
        {\tt \small  hxli@aibee.com}}
}
\begin{document}

\maketitle
\thispagestyle{empty}
\pagestyle{empty}

\begin{abstract}
We study the problem of learning a navigation policy for a robot to actively search for an object of interest in an indoor environment solely from its visual inputs. While scene-driven visual navigation has been widely studied, prior efforts on learning navigation policies for robots to find objects are limited. The problem is often more challenging than target scene finding as the target objects can be very small in the view and can be in an arbitrary pose. We approach the problem from an active perceiver perspective, and propose a novel framework that integrates a deep neural network based object recognition module and a deep reinforcement learning based action prediction mechanism. To validate our method, we conduct experiments on both a simulation dataset (AI2-THOR) and a real-world environment with a physical robot. We further propose a new decaying reward function to learn the control policy specific to the object searching task. Experimental results validate the efficacy of our method, which outperforms competing methods in both average trajectory length and success rate. 

\end{abstract}

\section{Introduction}

Developing an autonomous mobile robot which can reliably search, locate and reach an arbitrary object in an indoor environment is both fascinating and extremely challenging which motivates multi-disciplinary research ideas across robotics, computational perception, machine learning. In practice, a solution to this task will have a wide range of robotics applications, such as an assistant robot to search for survivors from an unknown disastrous environment for the first responders, or an elderly care-giving robot to locate and/or retrieve objects of interest for its clients. Solving this challenge has the potential to kick off the next phase of our human life style revolution that aims to increase people's living standard and enrich people's everyday life.
\begin{figure}[ht!]
\centering
\includegraphics[width=\columnwidth]{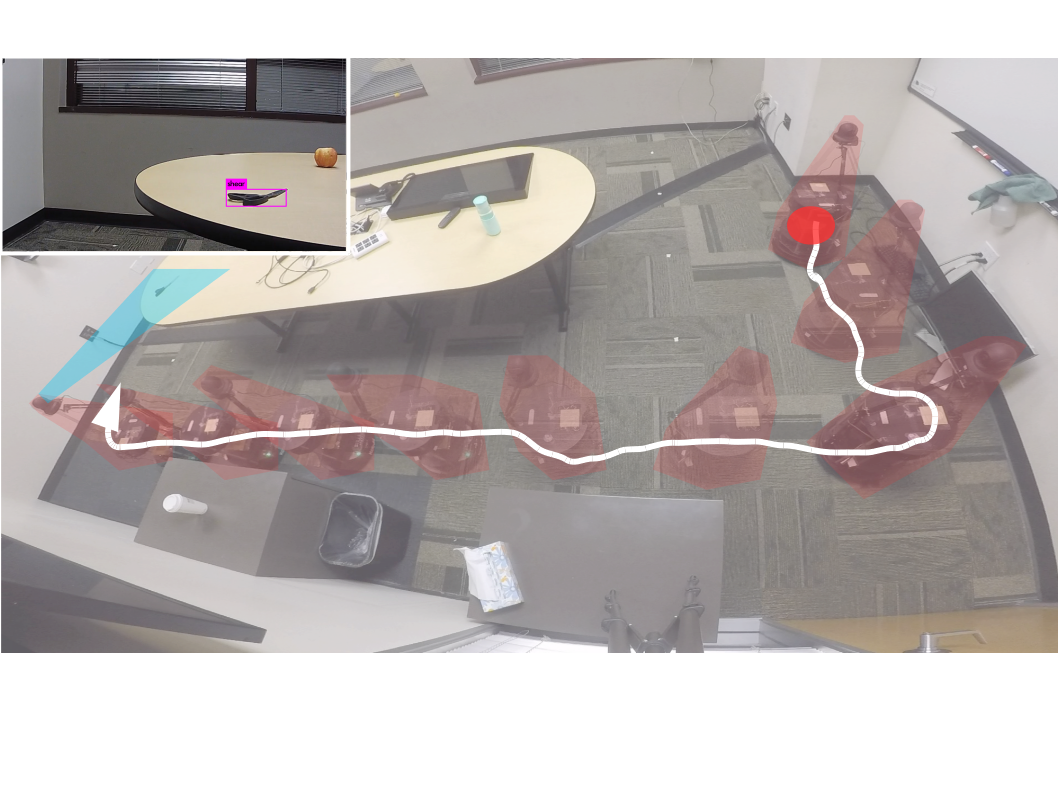}
\vspace{-4em}
\caption{An illustration of the ``Robot with vision that finds objects'' task. Red dot: the random initial location; White line and arrow: the generated robot trajectory from our turtle-bot experiment; Upper left image: the final view of the robot with the target object detection (the target object in this experiment is ``shear''.)}
\label{fig:todo}
\vspace{-1em}
\end{figure}

We fully acknowledge that studies approaching the problem have a long history. Tracing back to the 1970s and 1980s, when the concept coined as the ``active perception'' was widely explored, this ``robot with vision that finds object'' task was one of the major motivating tasks to show that ``vision is active'' \cite{tenenbaum1970accommodation}. As stated in a recent survey article \cite{bajcsy2018revisiting}, two primary aspects of ``active perception'' are 1) from intelligent control point of view, it is about intelligent control strategies applied to the perception process \cite{bajcsy1988active}, and 2) from computational perception point of view, it is about manipulating the perception constraints to improve the quality of the perceptual outputs \cite{aloimonos1988active}. These two aspects reflect two essential sub-tasks, object recognition and robot navigation, for enabling active object searching on mobile robots. 

With the coherent and organic unity in mind, from 1970s to early 2010s, we observed the two research areas, i.e. object recognition and robot navigation, are explored relatively independently and tremendous progress has been made in both tasks with novel computational models and algorithms. 



More recently, with the re-emergence of neural networks and deep learning, we witnessed significant breakthroughs in object recognition ~\cite{krizhevsky2012imagenet,redmon2016yolo9000}, and robotic motor control ～\cite{mnih2015human} with powerful convolutional neural networks and reinforcement learning algorithms. More importantly, the ``communication barrier'' between the two research areas starts to fade due to convergence of solutions in both tasks onto deep neural networks based models. This motivates us to explore a unified embodiment of yet another ``active perceiver'' and re-visit the task of ``robot with vision that finds object''. A special case in this task, ``robot with vision that finds a specific visual scene'', has been recently introduced in \cite{zhu2017target} based on joint modeling of image matching and navigation policy learning. Inspired by this work, in this paper, we attempt to move a step further to develop a joint model of object recognition and navigation policy learning to approach the general ``robot with vision that finds object'' task. Our object searching task is more challenging than the scene search task as the object can be very small, located in any location with an arbitrary pose.  

To study the aforementioned task, we collect a simulation dataset and a real-world dataset as benchmarks for training and evaluation. The simulation dataset is based on the recent AI2-THOR challenge platform \cite{ai2thor} which provides a simulation environment as a testbed. However, the challenge is designed for the aforementioned special case: ``robot with vision that finds a specific visual scene''. To further augment the dataset to test the general task of ``robot with vision that finds object'', we further supplement the benchmarking platform with objective annotations. A real-world dataset is collected with a real turtle-bot in a conference room at an office building with multiple, scattered objects. We intend to make both the augmented AI2-THOR data as well as the new real world dataset publicly available for follow-up research.



We summarize our contributions in three-fold: 1) we propose a new framework seamlessly integrating a deep neural network based object recognition module with a deep reinforcement learning based visual navigation module to learn for recognition-guided action policy learning; 2) we conduct experiments with a variety of reward functions and propose a new decaying reward accumulation scheme which yields the best performance; 3) we augment the AI2-THOR simulation platform with object bounding box labels and collect a new real world dataset to benchmark to evaluate our system and future research in this task.

\section{Related Work}

{\bf Object recognition:}
Object recognition and detection has been extensively studied during the past few years with deep learning approaches. One of the most popular methods are region proposal based, such as Fast R-CNN \cite{girshick2015fast} and Faster R-CNN \cite{ren2015faster}. These methods first generate multiple object proposals from the original image, then the features of these proposals are extracted to perform object classification and bounding box regression. YOLO \cite{redmon2016you} is the other kind of approach to do object detection. It adopts a single neural network to predict both bounding boxes and class probabilities from images without proposal generation. Some improvements to YOLO are also proposed by the authors in \cite{redmon2016yolo9000}. 
We also adopt a deep neural network based object recognition module as the guidance to our policy learning module, but different from those existing works, we feed an image of the target object as another network input (along with the current robot view) and train a class-agnostic object detector. We will later show the details of the module and explain why our module fits smoothly in the overall policy learning framework in Sec.~\ref{sec:or}.

{\bf Target-driven visual navigation:}
Among the approaches to vision-guided navigation policy learning, recently, \cite{zhu2017target} proposed a promising approach to tackle the problem of having robots match a given target scene solely based on visual inputs in indoor environments. They adopted deep reinforcement learning to learn the relationship between the current state (from camera input) and the actions that need to be taken to achieve the final goal (match a specific scene). In order to generalize this approach to different targets and scenes, they took the target state as the input and adopted scene-specific layers in the model. Building upon the first model, \cite{chenimitating} proposed several incremental extensions, whereas the extensions has limited contribution to better visual navigation performance.

{\bf SLAM supported object recognition: } \cite{pillai2015monocular} proposed a method to integrate object recognition, and simultaneous localization and mapping (SLAM) together into a unified framework, and experimentally showed that with the SLAM supported algorithm, the robot is able to recognize objects better. Instead of studying robot navigation to assist object recognition, we aim to leverage recognition as a guidance to help navigation, and object searching and reaching. 

{\bf Limitations of the previous work and our take: } The final target of the robot to find in both \cite{zhu2017target} and \cite{chenimitating} denotes a specific scene image, and the goal of the robot is to navigate to reach the location where the target image was taken. In such a scenario, the users must have access to the specific image, or have taken this image by themselves, which significantly limits the potential application of these methods. In this paper, we move a step further and tackle the task of learning policies for robots to allow object searching and reaching. Thus, the target image the user provides will only depict an instance of the object (i.e. an image of the object downloaded from web or an image of an object captured with a mobile camera), not exactly the object in the scene with the environment. Moreover, the robot is asked to detect the bounding box of the specified object in its viewpoint with a certain size to accomplish the task (Fig.~\ref{fig:todo} depicts the overall general task).

\section{Our Approach}
\subsection{Problem Formulation}
Our general idea is to learn action policies for an active agent (mobile robot) to locate a user-specified target object in indoor environments using only visual inputs (here, we assume a single stream image sequences from an on-board camera mounted in a robot). The target object is specified as an RGB image of the object which contains no contextual information. With the learned policies, when the user provides the robot an image of a target object, the robot is expected to take a relatively small number of steps to approach the target object from its random starting position. Moreover, the robot is expected to return a bounding box of the target object in its viewpoint (as shown in Fig~\ref{fig:todo}). 

Specifically, we propose a new deep reinforcement learning-based framework which integrates two basic modules: 1) an object recognition module trained to detect any given object in its viewpoint; and 2) a sequential decision making model guiding the robot to make action decisions at each time step and location to approach the target object. The input to the sequential decision making model includes the current visual observation captured by its camera, the target object image a user provides and the object location in the current observation (if any) that is detected by the object recognition module. We will describe the two modules respectively in the following sections.

\subsection{Object Recognition Module with Target Object Given}
\label{sec:or}

\begin{figure}[ht]
\centering
\includegraphics[width=\columnwidth]{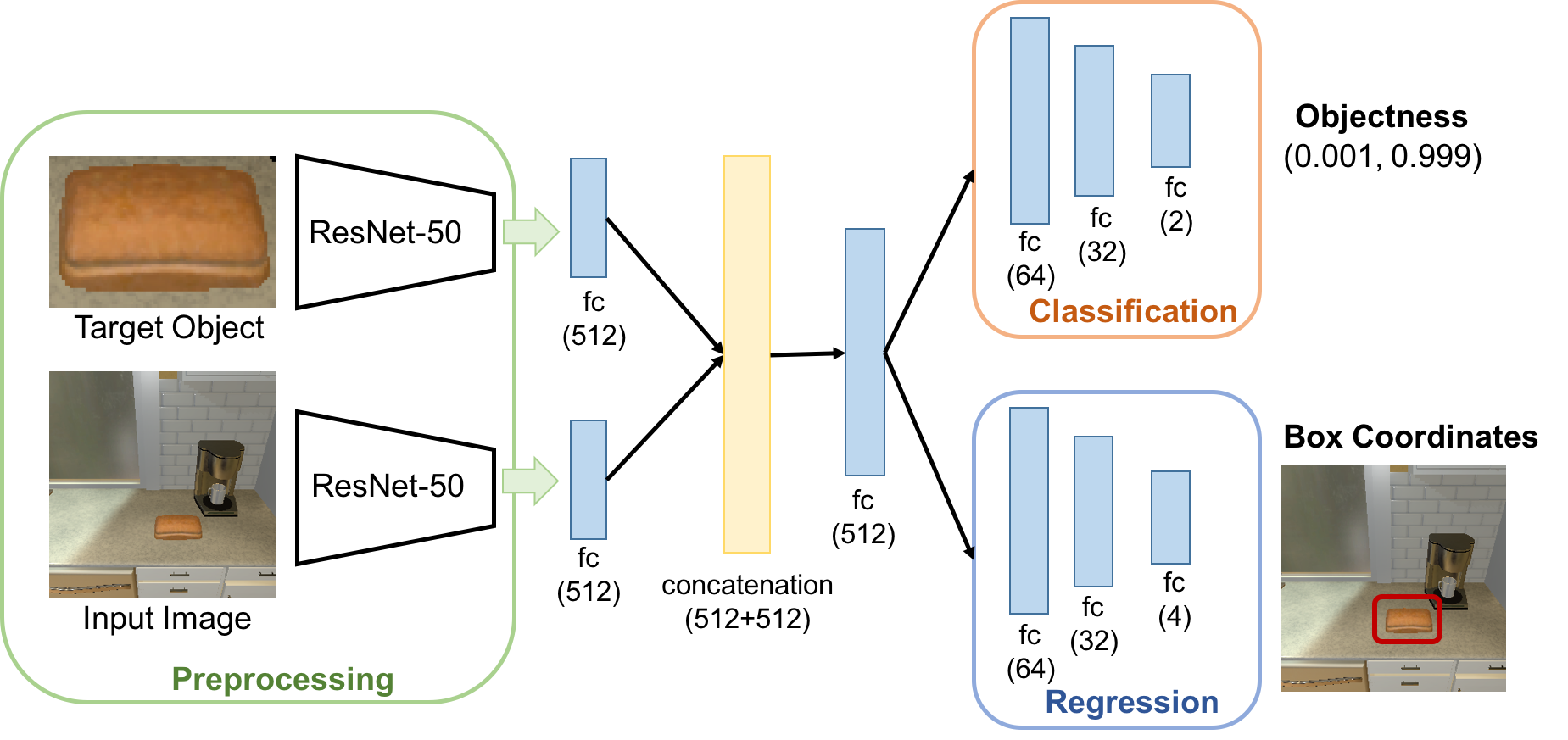}
\caption{The architecture of the object recognition network.}
\label{fig:object_detection}
\end{figure}

A standard object detection process typically consists of two steps: 1) detect the candidate regions of objects in an image; and 2) predict a class label to each region (a bounding box representation is typically used). However, under the active perceiver setting, the target object is given, and the object recognition module only needs to detect whether the specific object exists in the current range of view at any time step. Following this observation, we adopt a deep neural network that simply takes the target object image as the first input, along with the whole image of robot's current view as the second input, to predict the bounding box coordinates of the target object in the current view if there is one. Fig. \ref{fig:object_detection} shows an illustration of the proposed network architecture. To train such a network, we minimize the loss function as defined in Eq.~\ref{eq:loss}. We present the details in the following.


{\bf Network architecture:} As shown in Fig. \ref{fig:object_detection}, the module takes both the current view (one full image frame) and the target object image and feed them into a shared, ResNet-50 network \cite{he2016deep} to extract the $1024$ dimensional features at the output of the $res4f$ layer for both inputs. Here, we take the pre-trained ResNet model (trained on ImageNet) and fix them during our subsequent training and testing pipelines. A fully-connected layer is attached after each of the $1024$ dimensional feature inputs and projects each of them down into a $512$ dimensional vector. The two $512$ dimensional vectors are then concatenated and aggregated into a $512$ dimensional joint vector by an additional fully-connected layer. Finally, we feed the $512$ dimensional joint vector into a classification layer  and a regression layer, each of which includes another two fully-connected layers. The classification layer is designed to predict whether the target object appears in the input whole image or not, and the regression layer is designed to predict the $4$ parameters of the target object bounding box, i.e. the center coordinate $(x,y)$, the width $w$ and the height $h$ of the bounding box.

{\bf Loss function:} Eq.~\ref{eq:loss} shows the loss function we design for the object recognition module. 
\begin{equation}
\label{eq:loss}
\begin{aligned}
\mathbf{L} = &\sum_{i}{-\big[p_i^*log(p_i)+(1-p_i^*)log(1-p_i)\big]} \\
&+\lambda \sum_{i}{\mathbf{1}_i^{obj}\Vert\mathbf{b_i}-\mathbf{b_i^*}\Vert_2^2}
\end{aligned}
\end{equation}
The first term denotes the cross-entropy loss for binary classification, where $p_i^*$ is the ground-truth label of the $i$th image and its value equals to $1$ if the object appears in the image, and $0$ if it doesn't. $p_i$ is the predicted probability of the object appears in the image $i$ that is the output from the previous classification layer. The second term in the loss function is the $L_2$ loss between ground truth bounding box coordinates $b_i^*$ and that predicted coordinates $b_i$, where $b_i = (x_i, y_i, w_i, h_i)$ (likewise for $b_i^*$). $\mathbf{1}_i^{obj}$ is an indicator function indicates that if the object is in the image $i$. In this formulation, the regression loss will only be activated when the target object is detected in the current view. $\lambda$ is the weight factor that balances between these two losses. In practice we found a weight value of $0.5$ works well, so we fix it throughout all our reported experiments. The object recognition network is trained by minimizing the overall loss function with the standard stochastic gradient decent (SGD) optimization.

\subsection{Recognition-guided Action Policy Learning}
\label{rl}
With the object recognition module, the task is to make a decision on which action is the best to take given the current robot state. Here, we build upon a basic deep reinforcement learning based framework similar to \cite{zhu2017target}, in order to learn a mapping from the state space to the action space with the guidance from the recognition module. 

In general reinforcement learning setting, robot learns the optimal action policy through trail and error interactions with the environment. Specifically, at each time step, the robot takes an action to transit its current state to a new state, and receives a scalar reward as the feedback. The robot stops either it reaches the goal state or it runs out of the maximum number of steps. The optimal action policy is learned by maximizing the expected cumulative reward. Here, we adopt a deep neural network to approximate the policy function. We elaborate these key ingredients, as known as robot states, action space, network architecture and reward function under our active perceiver setting.

{\bf Robot states:}
Since we aim to handle an unknown indoor environment, the system is not able to access a global map and neither can it locate itself using odometers. The RGB image stream captured by the robot's camera is thus the only source of information that encodes the robot's current state. Since the goal of the robot is to find the target object, we define goal states as those ones captured when the robot's current view contains a bounding box of the target object. At the same time, the size of the bounding box needs to be larger than a predefined threshold to determine a success. In practice, we found the size of the fifth largest bounding box (among all the ground-truth bounding box instances) is a reasonable threshold, and it yields $5$ goal states (top $5$ images with the largest target object detected). 

{\bf Action space:}
To constrain the number of possible robot states and allow off-line batch learning, we consider the robot actions as the set of discrete movements in the physical searching space. In practice, the robot action space has eight different ones that can be categorized into: 1) Translation forward, backward, left or right with a fixed distance; 2) Tilt the mounted camera up or down with a fixed angle to adjust the current view; 3) Rotate left or right with $\frac{\pi}{2}$ angle to alter the frontal direction of the robot.

\begin{figure}[ht]
\centering
\includegraphics[width=\columnwidth]{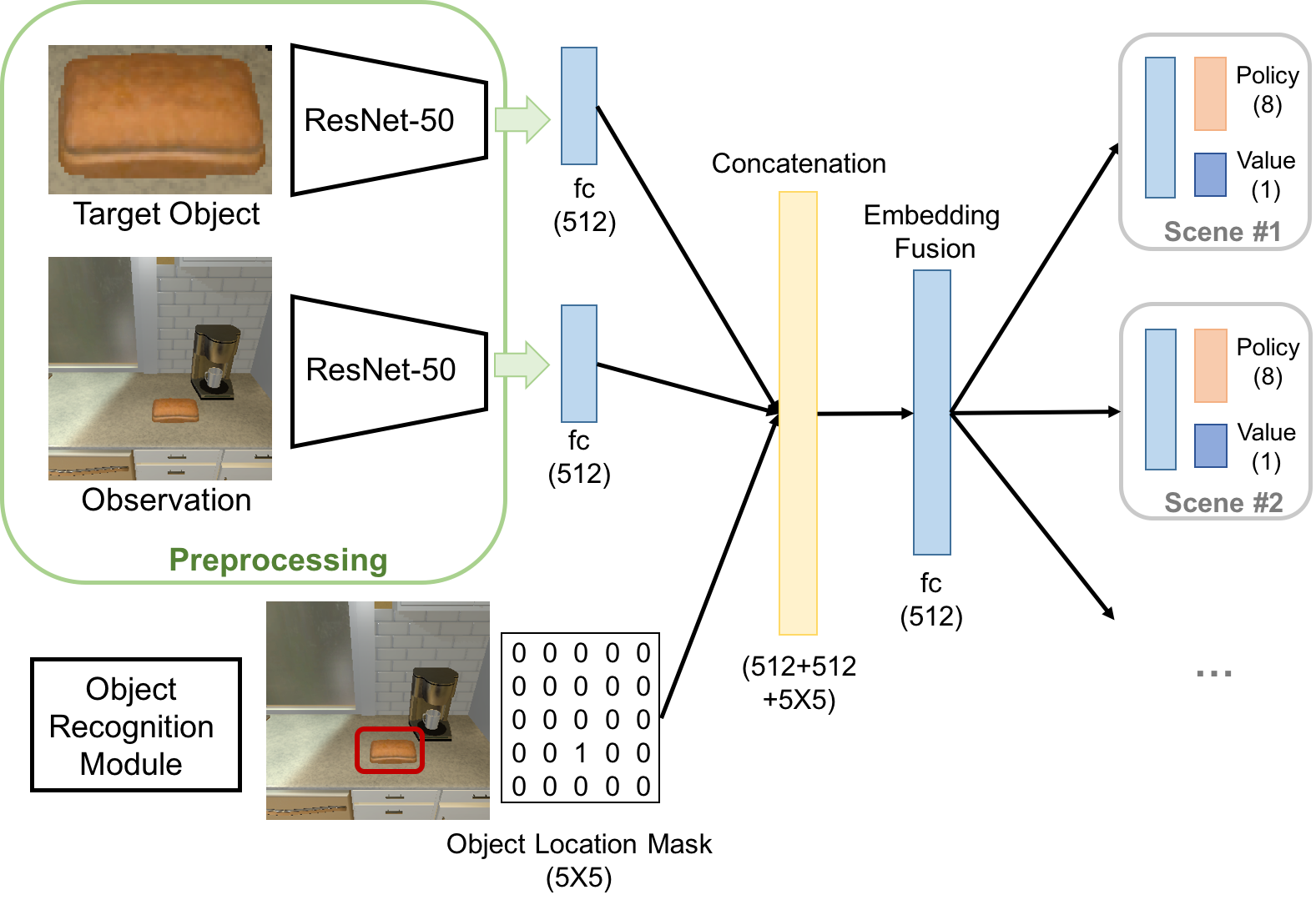}
\caption{The deep reinforcement learning architecture.}
\label{fig:rl}
\end{figure}

{\bf Network architecture:}
Fig. \ref{fig:rl} depicts the architecture of our proposed deep reinforcement learning-based model for the general active perceiver task. The model takes both the observed visual view and the target object image as the inputs. Then the scene-specific layer is used to predict the action policy for each scene. As suggested by \cite{zhu2017target}, such a network model has a good generalization ability across different targets and scenes. Our model does not share the weights between the two fully connected layers when fusing the target and input streams. One primary motivation for this is that under the general active perceiver setting,  the input image of the target object could come from very different domains from the object observed in the scene image so that it can't be projected to the same embedding space as the observation image. In addition, we further feed the object location information generated from the object recognition module into the embedding fusion layer. Here, the object location information is encoded as a $5$ by $5$ binary image that specify the object's location. We follow the same training protocol described from \cite{zhu2017target} to train this model.

{\bf Reward function:}
Most reinforcement learning problems adopt a straight-forward reward function design, which is to give a positive reward at goal states, and a fixed negative or zero rewards at all other intermediate states, (such as the one suggested in \cite{zhu2017target}). However, these reward functions usually require an extra care being taken on the exploit-exploration trade-off. More specifically, when the training process is trapped in a local optimum action policy, allowing more exploration to find more globally optimal policy usually leads to a slower convergence since more steps will be wasted by visiting many meaningless states. In the general active perceiver problem setting, due to the step limitation of the training episodes, the goal state may never be reached due to the low exploration rate, or the optimal policy may not converge when training is over. 

Here, we design a  more sophisticated reward function tailored for our task. It will improve the convergence rate in our problem greatly, but also can potentially generalize to other reinforcement learning tasks as it resists to local optima as shown in the experiments. 

For each state, we define the reward as the size (the area) of the detected bounding box that contains the target object, if there is one. For all states where the robot cannot detect the object, we set the reward to zero. Since there might exist many states where the robot can detect a bounding box with a smaller-than-the-threshold size, and these states tend to be next to each other in the exploration process, the robot can easily stuck between these states, since moving back and forth between them will always yield positive rewards. In order to encourage the robot to keep searching states with possibly larger detected bounding boxes, we further set reward as zero to a state,  if at this state the robot do detect a bounding box but the size of it is smaller than the bounding boxes that have been detected earlier in one episode. In other words, our proposed reward function keeps the records of the largest box size in the current episode and accumulates discounted rewards in an incremental way with respect to the records.  

More formally, consider $a_0, a_1, ..., a_s$ where $a_i(0\leq i \leq s) \geq 0$ are the areas of the bounding boxes the robot have seen during one episode, the total reward for this episode is set to be:  
\begin{equation}
\label{eq:rl}
\mathbf{Reward} = \gamma^{i_1}a_{i_1}+\gamma^{i_2}a_{i_2}+...+\gamma^{i_t}a_{i_t},
\end{equation}
where $a_{i_1} < a_{i_2} < ... < a_{i_t} (i_1 < i_2 < ... <i_t)$, and $\gamma$ denotes the discount factor for the penalty over time. The rationale supporting this design is that this reward function encourages more exploration only around these aforementioned and potential trapping states, instead of having a higher uniform exploration rate across the whole state space, regardless of whether these states being worth exploring or not. In this way, we can achieve both faster convergence and more meaningful exploration paths at the same time, and experimental results we observed validates the effectiveness of our setting.

\section{Experiments}
We evaluated our framework in both simulation and real environments. In the simulated environment,  we set up a variety of experiments with multiple target objects in four indoor scenes. We further implement our framework on a real mobile robot platform (a turtle-bot with a pan-tilt camera) and demonstrate its efficacy in a real indoor scene (a conference room) in finding the target objects. 

\subsection{Dataset}

In this section, we describe our datasets for our empirical evaluation in both simulation and real world scenario.

{\bf Simulation platform} is adopted from the THOR Challenge platform ~\footnote{http://vuchallenge.org/thor.html}. This platform provides $30$ photo-realistic indoor scenes ($15$ kitchens and $15$ living rooms) from AI2-THOR dataset \cite{ai2thor} for training ($20$ scenes) and validating ($10$ scenes) autonomous robotic systems to navigate and search for objects in these virtual environments. 
The available actions for the robot are predefined as $8$ discrete actions, namely $MoveAhead$, $MoveRight$, $MoveLeft$, $MoveBack$, $LookUp$, $LookDown$, $RotateRight$ and $RotateLeft$. Following the simulation setting, each virtual scene can be discretized into a set of images taken from each robot state, and the whole set of images characterize the overall state space of the robot. Moreover, the platform also provides a set of target images for each scene. These target images contain only objects without any background, which can well-support our experiments.

In order to train the object recognition module, we need to further augment the data with annotated object bounding boxes. Without the loss of generality, we select $4$ scenes randomly and for each scene we retrieve all images from every robot state. 
We further select $4$ objects that can be found in the scene and take their target images as the input target images to our system. Finally, we manually labeled the bounding box for each target object in each scene image, and treat them as the ground-truth bounding boxes in training our object recognition module. We adopted the standard cross-validation mechanism to mitigate the risk of over-fitting. 

\begin{figure}[ht!]
\centering
\begin{tabular}{c}
\subfloat[Sample testing images and corresponding locations on the map.]{\includegraphics[width=0.7\columnwidth]{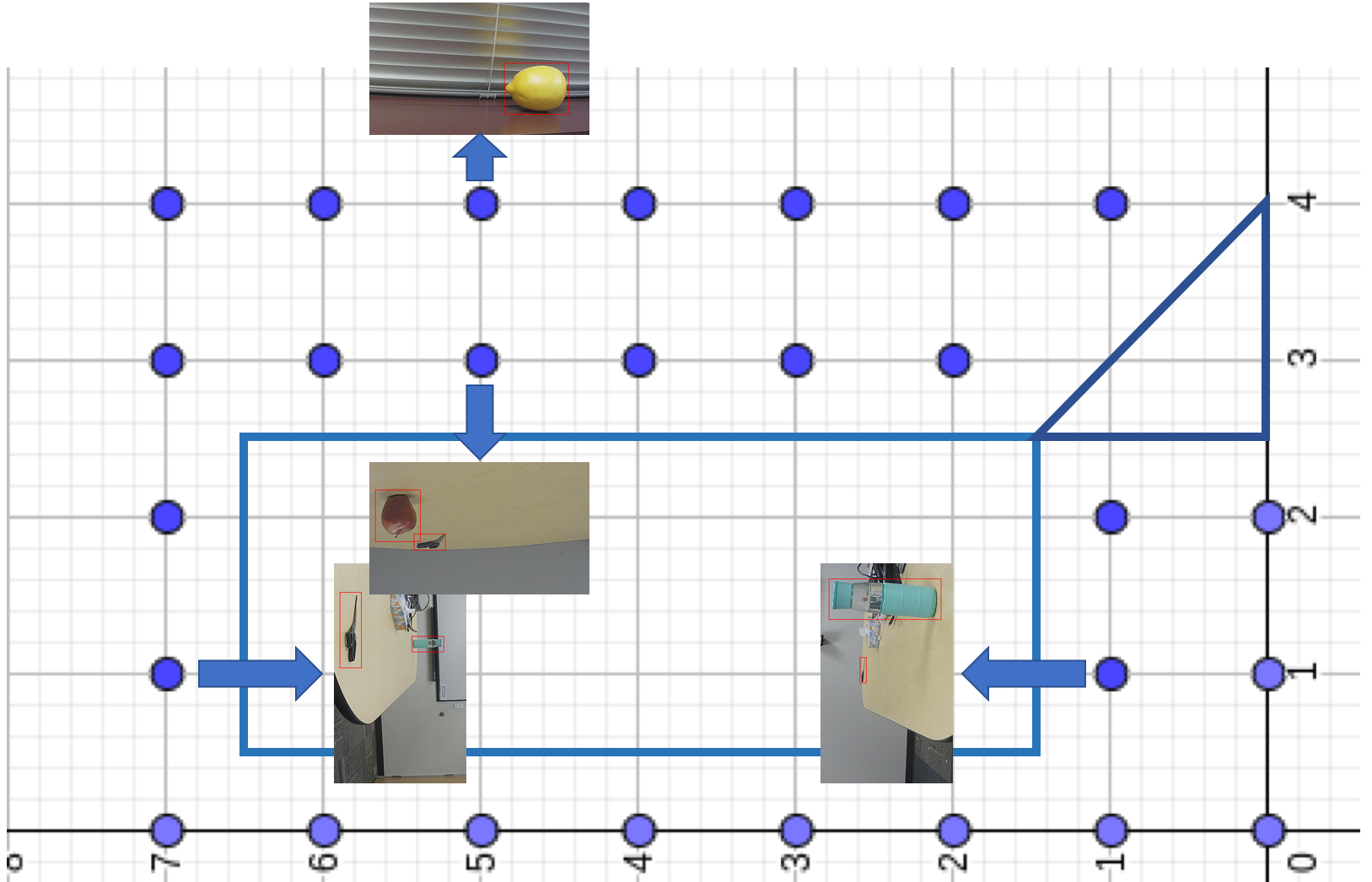}} \\
\subfloat[Sample training images taken from another room. ]{\includegraphics[width=0.9\columnwidth]{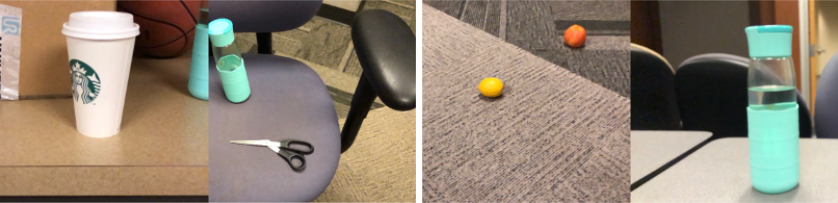}}
\end{tabular}
\caption{Sample testing and training images from our real world dataset. }
\label{fig:data}
\end{figure}

{\bf Real world scenario:}  we equipped a turtle-bot with a pan/tilt camera to conduct the experiment. The action set available for the turtle-bot consists of, $MoveAhead$, $MoveBack$, $RotateLeft$, $RotateRight$, $LookUp$, and $LookDown$ (pan/tilt camera). We discretized the scene space (which is a conference room) into $27$ locations. At each location, the robot takes $8$ RGB images with its pan/tilt camera (turn left/right, look up/down). This leads to a total of $216$ RGB images for our experiment, representing all possible states of the turtle-bot in this real world scenario. For the objects presented in the scene, we collected another group of $150$ images each for training the object recognition module. 
In testing, we set the turtle-bot at a random starting location in the conference room, and search for a given object with the trained model. Fig.~\ref{fig:data} shows a few sample testing and training data.

\subsection{Experimental Results}

\begin{table*}[ht]
\vspace{+1em}
\caption{The performance metrics of the different methods from the simulation and the real world experiments. }
\label{tbl:results}
\begin{center}
\vspace{-0.7em}
\begin{tabular}{|c|c|c|c|c|c|c|c|c|c|c|}
\hline
\multirow{4}{*}{Experiment Scenario} 
&\multicolumn{10}{c|}{Methods}  \\

\cline{2-11}

&\multicolumn{2}{c|}{\multirow{2}{*}{Random Walk}}
&\multicolumn{2}{c|}{\multirow{2}{*}{Reward Func. 1}}
&\multicolumn{2}{c|}{\multirow{2}{*}{Reward Func. 2}}
&\multicolumn{2}{c|}{Our Reward Func.}
&\multicolumn{2}{c|}{Our Reward Func.} \\

&\multicolumn{2}{c|}{}
&\multicolumn{2}{c|}{}
&\multicolumn{2}{c|}{}
&\multicolumn{2}{c|}{(high exploration)}
&\multicolumn{2}{c|}{(low exploration)}\\

\cline{2-11}
& Avg. len.&Suc. rate& Avg. len.&Suc. rate& Avg. len.&Suc. rate& Avg. len.&Suc. rate& Avg. len.& Suc. rate\\
\hline
$1$ object in $1$ scene &2050.3 & 60.0\% & 2880.2 & 50.0\% & -&0\% &1957.6 &80.0\% &52.9 &100\%\\
\hline
$4$ objects in $1$ scene &1911.6&72.5\%& $\sim$ & $\sim$& $\sim$  & $\sim$ &1643.2&92.5\%&30.1&75.0\%\\
\hline
$16$ objects in $4$ scenes &2057.6&84.3\%& $\sim$ & $\sim$ & $\sim$ & $\sim$ &1430.7&87.5\%&593.0&75.0\%\\
\hline
turtle-bot experiment &820.9&99.0\%& $\sim$ & $\sim$ & $\sim$ & $\sim$ & 176.8 & 100\% & 63.3 & 100\%\\

\hline
\end{tabular}
\vspace{-3em}
\end{center}
\end{table*}

\begin{table}[ht]
\caption{Model performance for each target object for the $4$ objects in $1$ scene setting.}
\vspace{-0.7em}
\label{tbl:len1scene}
\begin{center}
\begin{tabular}{|c|c|c|c|c|}
\hline
\multirow{2}{*}{Target Object} 
& \multicolumn{2}{c|}{High Exploration Rate} 
& \multicolumn{2}{c|}{Low Exploration Rate}   \\
\cline{2-5}

&Avg. len.&Suc. rate&Avg. len.&Suc. rate\\
\hline
Bread &2566.1&80\%& 52.9 & 100\% \\
\hline
Mug & 1426.8&100\%&36.8 & 100\%\\
\hline
Plate &1114.5&100\%&- & 0\% \\
\hline
Apple &2366.6&90\%&30.8  &100\% \\
\hline
\end{tabular}
\vspace{-3em}
\end{center}
\end{table}

\begin{figure*}[ht!]
\centering
\begin{tabular}{cccc}
\subfloat[Bread]{\includegraphics[width=0.47\columnwidth]{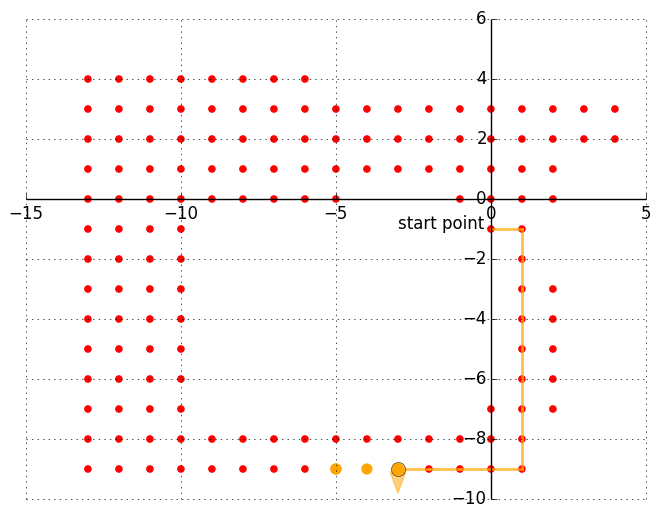}} &
\subfloat[Mug]{\includegraphics[width=0.47\columnwidth]{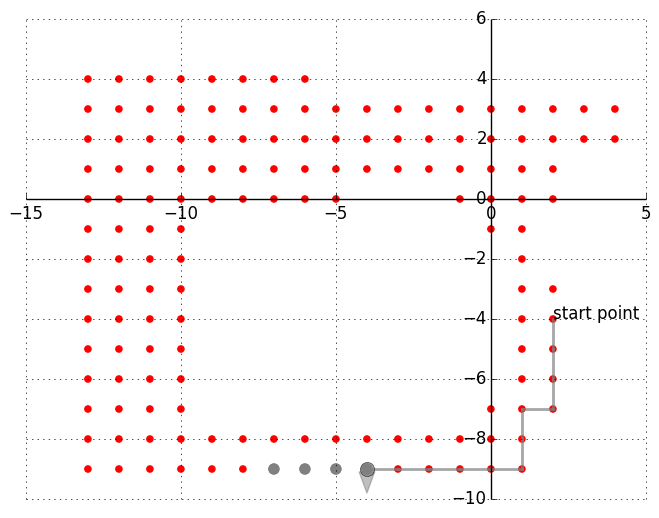}}  &
\subfloat[Apple]{\includegraphics[width=0.47\columnwidth]{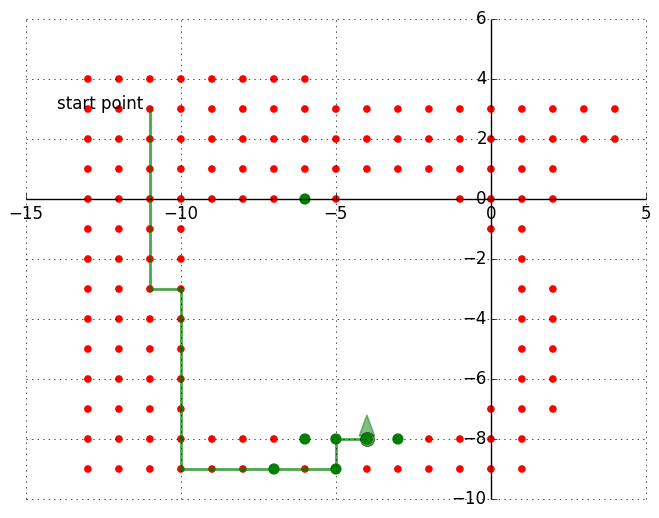}} &
\subfloat[Plate]{\includegraphics[width=0.47\columnwidth]{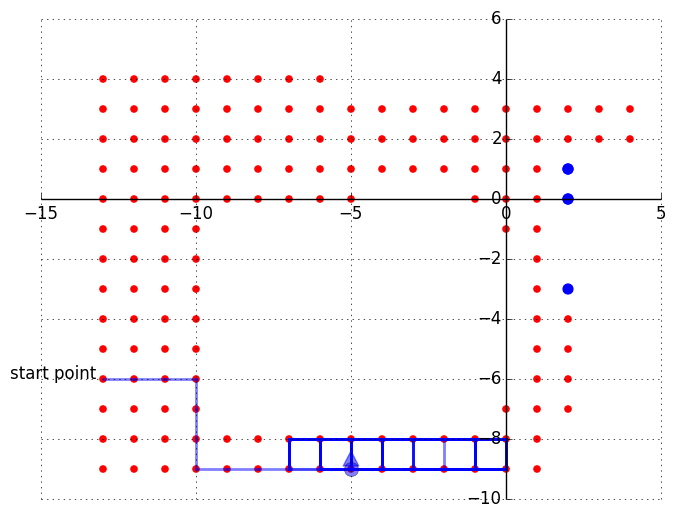}}
\end{tabular}
\caption{Robot's trajectory in finding each object. The smaller red dots denote the discrete scene space, and the bigger dots with different colors denote the goal states for different target objects. Here we only show the top-down view of the $2$ dimensional trajectories, omitting the orientations and the pan/tilt actions. }
\label{fig:trajectories}
\label{fig:trajectories}
\end{figure*}

TABLE \ref{tbl:results} reports the performance of different methods in finishing our tasks. We designed four tasks as following.

1) Locate $1$ object in $1$ simulated scene. In order to test if our proposed framework can successfully find the target object or not in one scene, we first conduct this experiment.

2) Locate $4$ objects in $1$ simulated scenes. To validate the generalization ability of our method to multiple objects and avoid over-fitting to one specific target object, we train one model to find any given one of $4$ different objects. 

3) Locate a total of $16$ objects in $4$ different simulated scenes. To verify that the models have the generalization ability across different scenes, we trained one model which has $4$ scene-specific branches as shown in Fig. \ref{fig:rl} to learn $4$ action policies for $4$ scenes. 

4) Turtle-bot experiment in a real world scenario. We conduct an experiment using a turtle-bot to search for an object in a real world scene (a conference room) to validate the real-world efficacy of our method.

For each experimental setting, we compare our method with the following baseline and variants to demonstrate the superiority of our method. Note that we don't compare our method with the classical search algorithms, such as $A^{*}$, depth-first or breadth-first search since we assume the global map of the environment is unknown. The robot will simply remain on the current state if the action it takes cause collision. This non-deterministic transition characteristic makes these deterministic algorithms unusable.

1) Random walk. In this baseline method, the robot randomly takes an action from its available action set at each state. We take this method as our baseline.

2) Reward function $1$. We adopt our model architecture but with a reward function different from the one defined in \cite{zhu2017target} in training. The reward function is defined to give a positive reward ($10$) at goal states, and a small negative reward ($-0.01$) for all the other states.

3) Reward function $2$. We trained our model with another intuitive reward function. We define the reward as the area of the object's bounding box at each state, no matter if the size of the current bounding box is smaller than the previous one. This means that if $a_0, a_1, ..., a_s$ where $a_i(0\leq i \leq s) \geq 0$ are the areas of bounding boxes the robot have seen during one episode, the total reward for this episode is $a_0+\gamma a_1 + ... + \gamma^{s} a_s$ and $\gamma$ is the discount factor.

4) Our reward function with high exploration rate. We use our reward function defined in Eq.~\ref{eq:rl} to train our model, with a relatively high exploration rate.

5) Our reward function with a low exploration rate. We reduce the exploration rate to train our model without any other change compared to method 4).

After the model is trained, it takes about 0.5 seconds (on an Nvidia GeForce GTX 1080 Ti machine) to feed forward our object recognition network and deep reinforcement learning network to generate an action at each state.
We report the performance of all methods with two metrics, namely the average number of actions that need to be taken to find the target object (also known as the average trajectory length), and the success rate of the robot finally finds the target object. For a fair comparison, for each target object, we randomly initialize the robot's starting position and run each method for $10$ episodes. An episode ends when the robot finds the target object or it has already taken $5000$ steps (and it claims a failure). Only when the robot finds the target object successfully, we count the episode as a successful trail, and the corresponding trajectory length will then contribute to the average trajectory length metric. ``-'' indicates the model does not converge or during testing the robot is trapped in sub-optimal states for good. Here we want to mention a caveat: because both the reward function 1 and 2 do not perform well even on the $1$ object $1$ scene scenario and due to the limited computing resource we have, we did not test them for the other three scenarios (we use ``$\sim$'' in the table).


\begin{figure}[ht!]
\centering
\includegraphics[width=0.85\columnwidth]{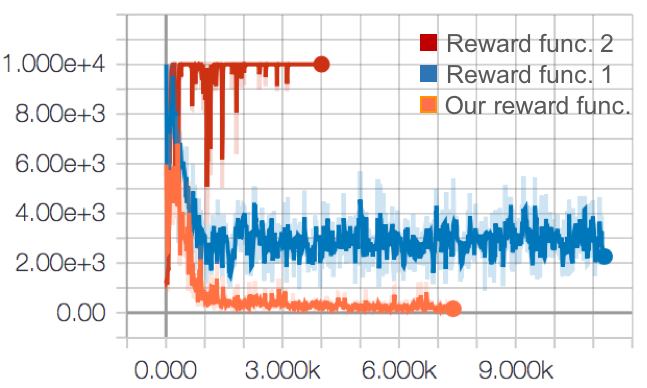}
\caption{Illustration of the average trajectory length profiles. Using the reward function 1 from \cite{zhu2017target} and reward function 2 the model is not able to converge.}
\label{fig:reward}
\vspace{-1em}
\end{figure}

We observed that: 1) trained model based on our proposed reward function outperforms other baseline methods with a significant margin (less steps to find the target object with a higher success rate; 2) our experiments show that reward function 1 and 2's performances are worse than random walk, or can not converge (see Fig.~\ref{fig:reward}),  which indicates they are not suitable for the general ``robot with vision that finds object'' task;  3) model trained with multiple objects and multiple scenes outperforms the model trained on one object with one scene, it partially reflects the generalization capability of the system; and  4) our reward function is sensitive to the exploration rate, which we will discuss in details in the following section. 

\subsection{Analysis and Discussion}
\vspace{-1.2em}
\begin{figure}[ht!]
\centering
\begin{tabular}{ccc}
\subfloat[Initial Entropy]{\includegraphics[width=0.3\columnwidth]{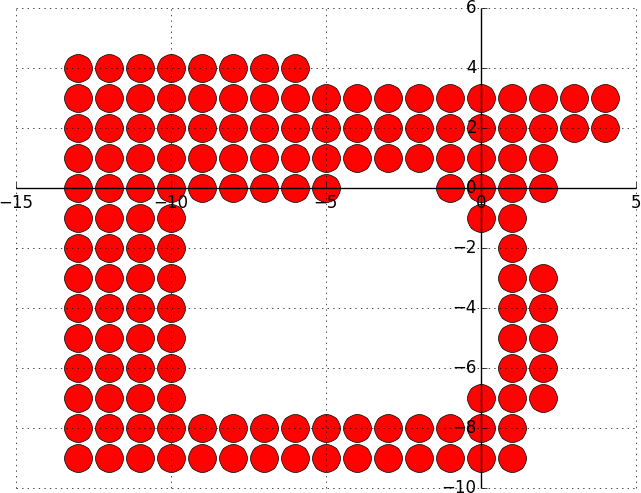}}
&\subfloat[After 10000 iterations of training]{\includegraphics[width=0.3\columnwidth]{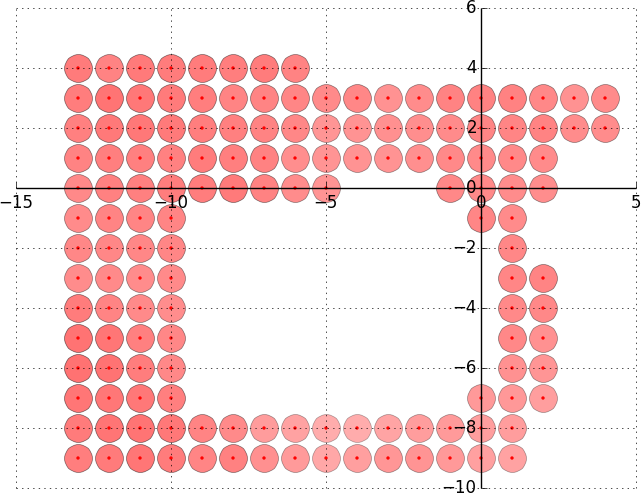}}
&\subfloat[After 30000 iterations of training]{\includegraphics[width=0.3\columnwidth]{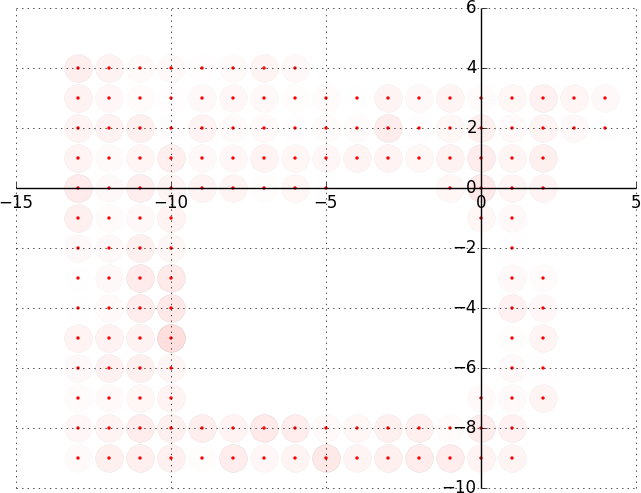}}
\end{tabular}
\caption{Illustration of the evolution of the average entropy over the action policy at each $2$-d position for one testing scene (a simulated kitchen environment). Higher entropy with darker color coded corresponds to more random action policy, and vice versa. }
\label{fig:entropy}
\end{figure}

{\bf Generalizing over multiple objects and the role of the exploration rate: }we further analyze our method more closely under the scenario of finding $4$ objects in $1$ scene. In this case, our deep reinforcement learning model as shown in Fig.\ref{fig:rl} needs only one scene-specific layer. Following the training protocol, we assign $4$ threads, each of which takes one of the four target object images as an input and learns the action policy to find the object. All threads keep a copy of the global network and update the weights of the global network per each episode. 

We trained this model using our reward function with either a high or a low exploration rate. In practice, we add a negative entropy of the predicted action policy to the loss function of our deep reinforcement learning model to encourage exploration. We control the exploration rate by changing the weight of this entropy item. Here, we set a weight equals to $0.1$ as a high exploration rate, and $0.01$ as a low exploration rate. During testing we observed that the model with high exploration rate has long average trajectory length but high success rate, while the low exploration rate model has much shorter average trajectory, however the success rate is relatively low.

To further analyze the performance, we list the two metrics achieved for each object in TABLE \ref{tbl:len1scene}. It shows that with the model of a high exploration rate, it is able to find all objects from various random starting points, although the average trajectory taken is longer. While under the low exploration setting, the robot can find the three objects within almost optimal amount of steps, but it fails to find the object ``plate'' for all trails. To further discuss the observed experimental results, we explore the scene and depict robot's trajectory generated by the model with low exploration rate in Fig. \ref{fig:trajectories}. We find that the goal states of the other $3$ objects are actually close to each other in this scenario. Such a case may provide the robot strong prior knowledge during the training process that the target object is among these locations. As a result, the robot would first go to check these locations, and with low exploration rate, the robot would have a high chance to stuck there. This observation indicates a trade-off between the average trajectory length and the success rate, and setting a proper exploration rate becomes critical.

{\bf Action policy learning: }
To further illustrate that our action policy learning through the proposed system and reward function design converges well, we show the average entropy of the action policy at each $2$-d position in Fig.~\ref{fig:entropy}. Here we calculate the entropy as follows: $\mathbf{Entropy(\pi)} = -\sum_{i=1}^{n}{\pi(i) log \pi(i)}$,
where $\pi$ is the action policy (a belief distribution over all possible actions ($n$ here) to take given the target object). After around $30000$ iterations of training, the figures show that our model converges well.

\section{Conclusion and Future Work}

In this paper, we presented an active object perceiver system to enable ``robot with vision that finds object'' that features two novel improvements from previous ones: 1) we proposed and implemented a object recognition-guided policy learning mechanism through an integration of two deep neural networks based architectures; 2) we put forward a novel decaying reward function for the deep reinforcement learning part, and augment the public available dataset with new object-hood annotations.  Experiments conducted on both public AI2-THOR \cite{ai2thor} platform and a new indoor environment dataset (a conference room), in simulation and on a physical robot executing the challenging object finding task validated the proposed approach. 

The study presented in this paper also opens several avenues for future studies. First, an active perceiver also relies on its memory unit to decide whether a state has been visited. In such a case,  further research on integrating an explicit memory unit is needed. Moreover, humans bring prior knowledge about the world to enable efficient decision making. Especially for this object finding task, how to incorporate both action policy learning through exploration with structured forms of common-sense knowledge (such as cup is more likely to be found on a table than on the floor)  requires an explicit knowledge distillation mechanism. %

 \textbf{Acknowledgments. }This work is partially supported by NSF CAREER IIS-1750082 and a gift from Adobe. We acknowledge NVIDIA for the donation of GPUs. 

\bibliography{references}
\bibliographystyle{IEEEtran}

\end{document}